\documentclass[10pt,twocolumn,letterpaper]{article}
\usepackage[tables,usenames,dvipsnames]{xcolor}
\usepackage{3dv}
\usepackage{times}

\usepackage{adjustbox}
\usepackage{amsmath}
\usepackage{amssymb}
\usepackage{amsthm}
\usepackage{array}
\usepackage{booktabs}
\usepackage{cuted}
\usepackage{float}
\usepackage{graphicx}
\usepackage{multirow}
\usepackage{pifont}
\usepackage{soul}
\usepackage{subcaption}
\usepackage{tikz,graphicx}
\usepackage[pagebackref=true,breaklinks=true,colorlinks,bookmarks=false]{hyperref}
\usepackage[noabbrev,capitalize]{cleveref}
\usepackage[square,numbers,sort]{natbib}

\threedvfinalcopy{}

\newcommand{\method}{Neural Feature Fusion Fields\xspace}
\newcommand{\acronym}{N3F\xspace}

\makeatletter
\renewcommand{\paragraph}{%
  \@startsection{paragraph}{4}%
  {\z@}{0.25em}{-1em}%
  {\normalfont\normalsize\bfseries}%
}
\makeatother

\newcommand{\improve}[1]{\textcolor{ForestGreen}{#1}}

\begin{document}

\title{\method:\\
3D Distillation
of Self-Supervised 2D Image Representations}

\author{
Vadim Tschernezki$^{1,2}$~~~%
Iro Laina$^1$~~~%
Diane Larlus$^2$~~~%
Andrea Vedaldi$^1$ \\
\centering
\begin{minipage}[t]{.4\textwidth}
\vspace{0pt}
\centering
\small{%
~\\ ${}^1$ Visual Geometry Group, University of Oxford\\}
{\tt\small \{vadim,iro,vedaldi\}@robots.ox.ac.uk}
\end{minipage}
\begin{minipage}[t]{.4\textwidth}
\vspace{0pt}
\centering
\small{
~\\ ${}^2$ NAVER LABS Europe\\}
{\tt\small diane.larlus@naverlabs.com}
\end{minipage}
}

\twocolumn[{%
\renewcommand\twocolumn[1][]{#1}%
\maketitle
\thispagestyle{empty}
\vspace{-1.5em}
\begin{center}
  \centering
  \captionsetup{type=figure}
      \includegraphics[width=\linewidth]{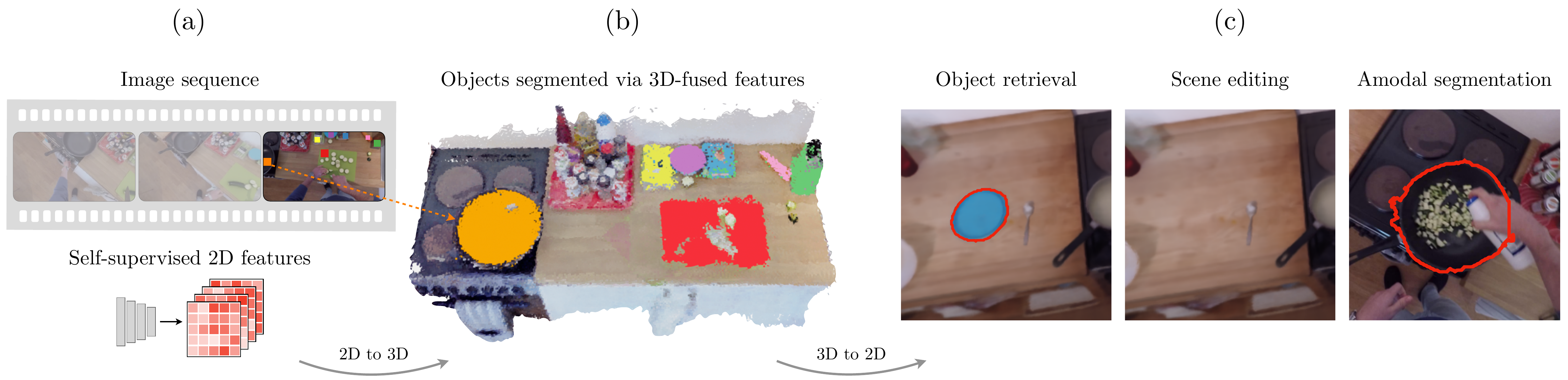}
\captionof{figure}{
\textbf{\method:}~%
(a) Given a collection of 2D images and a dense self-supervised feature extractor such as DINO, our method distills the features into a 3D representation via neural rendering. This allows to operate within the learned scene representation through 2D inputs. (b) For example, prompted with the features of a 2D region (any of the colored patches in (a)), our method segments the corresponding object in 3D as shown in the point cloud.
(c) We can also render the scene representation and solve image-level tasks such as object retrieval, scene editing, or amodal segmentation.}%
\label{fig:att}%
\end{center}%
\vspace{1.5em}
}]

\begin{abstract}
\vspace{-1em}
We present \method (\acronym), a method that improves dense 2D image feature extractors when the latter are applied to the analysis of multiple images reconstructible as a 3D scene.
Given an image feature extractor, for example pre-trained using self-supervision, \acronym uses it as a teacher to learn a student network defined in 3D space.
The 3D student network is similar to a neural radiance field that distills said features and can be trained with the usual differentiable rendering machinery.
As a consequence, \acronym is readily applicable to most neural rendering formulations, including vanilla NeRF and its extensions to complex dynamic scenes.
We show that our method not only enables semantic understanding in the context of scene-specific neural fields without the use of manual labels, but also consistently improves over the self-supervised 2D baselines.
This is demonstrated by considering various tasks, such as 2D object retrieval, 3D segmentation, and scene editing, in diverse sequences, including long egocentric videos in the EPIC-KITCHENS benchmark.
Project page: {\small\url{https://www.robots.ox.ac.uk/~vadim/n3f/}}
\end{abstract}

\vspace{-3em}
\section{Introduction}%
\label{s:introduction}

With the advent of machine learning two decades ago, computer vision has shifted its focus from 3D reconstruction to interpreting images mostly as 2D patterns.
Recently, however, methods such as NeRF~\cite{mildenhall20nerf:}
have shown that even 3D reconstruction can be cast effectively as a learning problem.
However, such methods are often optimized per scene, resulting in low-level representations of appearance and geometry that do not capture high-level semantics.
It is thus compelling to rethink how 3D reconstruction can be integrated with semantic analysis at image level to obtain more holistic scene representations.

In this paper, we consider a simple, general and effective approach for achieving such an integration, which we call \emph{\method} (\acronym; see~\cref{fig:att}).
The key idea is to map semantic features, initially learned and computed in 2D image space, to equivalent features defined in 3D space.
Forward mapping from 3D space to 2D images uses the same neural rendering equations as for view synthesis in prior work.
From this, backpropagation can move features from the 2D images back to the 3D model.

Recent methods such as Semantic NeRF~\cite{zhi21in-place} and Panoptic NeRF~\cite{fu2022panoptic,kundu22panoptic} have described a similar process for semantic and instance segmentation via label transfer.
Our intuition is that fusion does not need to be limited to image \emph{labels}, but can be extended to any image \emph{features}.
\acronym follows a student-teacher setup, where features of a 2D teacher network are distilled into a 3D student network.
We show that distilling the features alongside 3D geometry via neural rendering significantly boosts their consistency, viewpoint independence, and occlusion awareness.
As a result, the student ``surpasses the teacher'' in understanding a particular scene and improves tasks such as object retrieval and segmentation from visual queries.

As a particularly compelling case of this idea, we consider starting from a \emph{self-supervised} feature extractor.
Recent work~\cite{caron21emerging,amir2021deep,melas-kyriazi22deep,tumanyan22splicing} has shown that self-supervised features can be used to identify object categories, parts, and their correspondences in an ``open-world'' setting, \ie~without committing to a specific set of labels and without collecting annotations for them.
This is of particular relevance for emerging applications such as egocentric video understanding, where image understanding must work in user-specific and constantly evolving scenarios.

Specifically, we consider two scenarios of increasing difficulty.
First, we validate our contribution on simple \emph{static} scenes with only one or a few objects of interest, and combine \acronym with the vanilla NeRF model~\cite{mildenhall20nerf:}.
Second, we consider the more challenging scenario of egocentric videos which include static but also \emph{dynamic} components.
We adopt the same setting as NeuralDiff~\cite{tschernezki21neuraldiff} and consider videos from EPIC-KITCHENS~\cite{damen18epic} which contain long sequences of actors cooking in first-person view.

Given these diverse sets of videos, we use object retrieval (\eg, one-shot recognition) as a proxy to evaluate the quality of the fused image features.
Considering an object instance in a single frame, we use it to pool the features from \acronym, and then retrieve other occurrences of the same object in the rest of the video, despite severe viewpoint changes or occlusions.
We show that, while 2D features already perform well for this task, \acronym systematically boosts performance by a large margin.
This observation is consistent for several self-supervised and supervised features.
We illustrate other benefits of such an integrated model by also showing results for the tasks of
3D object segmentation, amodal segmentation, and scene editing.

\section{Related Work}%
\label{s:related}

We summarize relevant background work in feature extraction, reconstruction and neural rendering.

\paragraph{Self-supervised visual features.}

While \acronym can work on top of any 2D dense image features, including recent ones based on Vision Transformers (ViT)~\cite{dosovitskiy2020vit} and variants~\cite{touvron2020deit,yuan2021tokens, zhou2021deepvit, chen2021crossvit, touvron2021cait, bao2022beit, heo2021rethinking, graham2021levit, wu2021cvt, liu2021swin}, of particular interest are self-supervised versions such as~\cite{caron21emerging, chen2021empirical, li2022efficient, he2022masked} as they are more generically applicable and can benefit more from the consistency induced by \acronym.
Caron~\etal~\cite{caron21emerging} observed that their method (DINO), trained with self-distillation, learns better localized representations, which can be used to segment salient objects \emph{without any labels}.
Subsequently these features have been used for unsupervised object localization~\cite{melas-kyriazi22deep,LOST,wang2022tokencut}, semantic segmentation~\cite{melas-kyriazi22deep,ziegler2022self,hamilton2022unsupervised}, part segmentation~\cite{amir2021deep, choudhury2021unsupervised} and point correspondences~\cite{amir2021deep}.

\paragraph{Neural rendering.}
Using implicit representations of geometry in vision dates back at least to level-set methods~\cite{osher04level}.
Recently, authors have proposed to represent implicit functions with deep neural networks for the representation of geometry~\cite{park19deepsdf} and radiance fields~\cite{sitzmann19scene}, fitting the latter to 2D images via differentiable rendering.
Neural Radiance Fields (NeRFs)~\cite{mildenhall20nerf:} have popularized such ideas by applying them in a powerful manner to a comparatively simple setting:
novel view synthesis from a collection of images of a single static scene.
They combine radiance fields with internal learning~\cite{ulyanov18deep} and various architectural improvements such as positional encoding to obtain excellent scene reconstructions.
For a comprehensive overview of recent trends in this field see~\cite{tewari2022advances,neuralfields}.

Among countless extensions of NeRF, of particular interest for our applications are versions tackling dynamic scenes.
For example, NSFF~\cite{li21neural} models scenes through time-dependent flow fields, which enable novel view synthesis in space \emph{and} time; other methods achieve a similar effect by introducing canonical models~\cite{park2020nerfies,pumarola2021d,chen2021animatable,peng2021animatable} or space warping~\cite{tretschk2021nonrigid}.
Used here, NeuralDiff~\cite{tschernezki21neuraldiff} extends the standard NeRF reconstruction of the static part of a scene with two dynamic components, one for transient objects (foreground), and one for the actor in egocentric videos.

\paragraph{Semantic and object-centric neural rendering.}

Radiance fields provide low-level representations of geometry and radiance and lack a higher-level (\eg, semantic or object-centric) understanding of the scene.
Several works employ neural rendering to decompose multi-view or dynamic scenes into background and foreground components~\cite{yuan2021star,ost2021neural,tschernezki21neuraldiff,yu2022unsupervised,sharma2022seeing,ren2022neural}, while others focus on modeling the scenes as compositions of objects~\cite{yang2021learning, wu2022object, niemeyer2021giraffe, guo2020object,lazova2022control,xie2021fig}.
Some authors propose to combine radiance fields with image-language models (\eg, CLIP~\cite{radford2021learning}) to achieve semantically aware synthesis~\cite{jain2021putting,wang2022clip}.

More related to our work, however, are methods that extend radiance fields to also \emph{predict} semantics~\cite{zhi21in-place, kohli2020semantic, vora2022nesf, zhi2021ilabel}.
For example, Semantic-NeRF~\cite{zhi21in-place} has done so by using differentiable rendering to achieve multi-view semantic fusion of 2D labels akin to~\cite{mccormac2017semanticfusion,hermans2014dense,vineet2015incremental}.
NeSF~\cite{vora2022nesf} focuses instead on inferring semantics jointly across various scenes, using density fields as input to a 3D segmentation model.
However, it is only demonstrated on synthetic scenes with a limited number of categories and shapes.
Panoptic (\ie, semantic and instance) labels, have also been considered:
{}\cite{fu2022panoptic} uses NeRF as a means to integrate coarse 3D and noisy 2D labels and render refined 2D panoptic maps, while~\cite{kundu22panoptic} proposes an object-aware approach that can handle dynamic scenes, where each 3D instance is modeled by a separate MLP\@.
All of these methods use semantic labels to train their models and in particular the latter two require 3D labels.
Instead, our approach builds on self-supervised features and can yield a 3D-consistent semantic segmentation of static and dynamic scenes without \emph{any} labels.

The most related work is the concurrent paper by Kobayashi~\etal~\cite{kobayashi2022decomposing} who propose to fuse features in the same manner as we do;
they mainly differ in the example applications, including the use of multiple modalities, such as text, image patches and point-and-click seeds, to generate queries for segmentation and, in particular, scene editing.

\paragraph{Feature distillation.}

The motivation behind distillation originates from the task of compressing, or ``distilling'', the knowledge of large, complex model ensembles into smaller models, while preserving their performance~\cite{bucilua2006model}.
Hinton~\etal~\cite{hinton2015distilling} have shown that the performance of a distilled (student) model can even improve over the performance of the original model or model ensemble (teacher) when following the teacher-student paradigm.
Many methods have since then proposed to use this paradigm on features, tackling the task of feature distillation~\cite{romero2015fitnets, yim2017knowledge, park2019relational, tian2020contrastive, passalis2018learning,heo2019comprehensive,zagoruyko2017paying}.
While \acronym also makes use of the teacher-student paradigm, it differs in that the output of a 2D teacher network is distilled into a student network that implements a 3D feature field, resulting in different domains of the student (2D images) and teacher (3D points).

\section{Method}%
\label{s:method}

We first describe \method (\acronym) for generic (\cref{s:neuralfusion}) and advanced (\cref{s:advanced}) neural rendering models, and then introduce a number of applications (\cref{s:applications}) which we use to demonstrate its benefits.

\begin{figure*}
\centering
\includegraphics[width=1\linewidth]{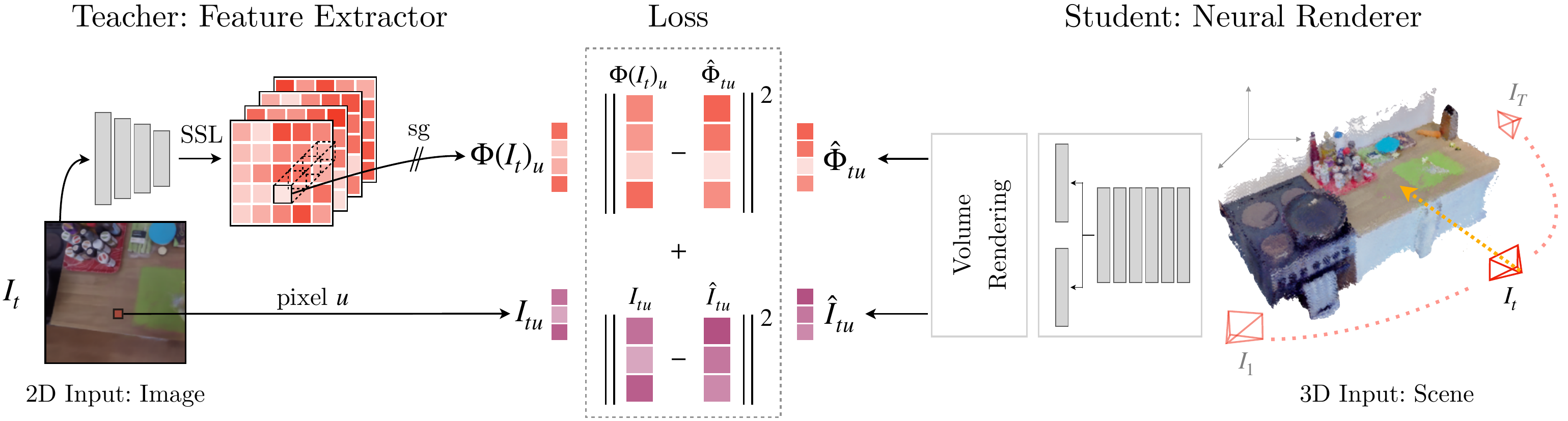}
\caption{
\textbf{Overview of our approach.}
\acronym follows a student-teacher setting where features computed from individual images are distilled into a 3D student network. The student network extends NeRF-like models such that a ray from a selected view is mapped to a color value $\hat{I}_{tu}$ and a corresponding feature vector $\hat{\Phi}_{tu}$ through volumetric rendering.
The teacher network, which is learned with self-supervision (SSL), predicts the 2D image features $\Phi(I_t)_u$ to be distilled.
The student is trained to optimize both image and feature reconstruction objectives, whereas the teacher is not trained further (stop gradient or `sg').
While the student network solely learns from 2D features, the resulting representation can operate either in 2D or in 3D.}%
\label{fig:method}
\end{figure*}

\subsection{\method}%
\label{s:neuralfusion}

Let $I\in\mathbb{R}^{3\times H\times W}$ be an input image defined on the lattice $\Omega = \{1,\dots,H\} \times \{1,\dots,W\}$ and let $\Phi$ be a feature extractor, \ie a function mapping the image $I$ to a representation $\Phi(I)$.
We assume the representation is in the form of a vector field $\mathbb{R}^{C\times H\times W}$, which is in itself an image with $C$ feature channels.
Example features include dense SIFT features~\cite{lowe2004distinctive}, convolutional networks~\cite{chen2017deeplab} and visual transformers~\cite{dosovitskiy2020vit}.
Furthermore, these features can be handcrafted, supervised or unsupervised.

Now suppose that the image is part of a collection $\{I_t\}_{1\leq t\leq T}$ and that the camera parameters are given, so that the projection function $\pi_t$ from world coordinates $X\in\mathbb{R}^3$ to image coordinates $u = \pi_t(X) \in \mathbb{R}^2$ is known.
A \emph{neural radiance field} is a pair of functions $(\sigma, c)$ mapping 3D points $X\in\mathbb{R}^3$ to occupancy values $\sigma(X)\in\mathbb{R}_+$ and to colors $c(X)\in\mathbb{R}^3$ respectively. In practice, color also depends on the viewing direction $d \in \mathbb{R}^3$, but our notations omit this dependency for brevity.
The neural rendering equation reconstructs the color $I_{tu}$ of pixel $u$ as
\begin{equation}
\label{e:nerf}
\hat I_{tu}
=
\int_0^\infty
c(X_{tu}(r)) \sigma(X_{tu}(r))
e^{- \int_0^r \sigma(X_{tu}(q))\,dq}
\,dr
\end{equation}
where ${\{X_{tu}(r)\}}_{r > 0}$ are points along the ray from the camera center through pixel $u$ in image $I_t$.

The idea of neural rendering is to \emph{learn} the functions $\sigma$ and $c$
given only the images $I_t$ and the camera poses $\pi_t$ as input.
This is done by minimizing the image reconstruction loss
$
\sum_{t} \| \hat I_t - I_t \|^2
$
for all images in the sequence with respect to the parameters of the models $\sigma$ and $c$.

In \acronym, we propose to generalize this model by also reconstructing \emph{feature} images $\Phi(I_t)$ instead of just color images $I_t$. For this, we also minimize the loss:
\begin{equation}
\label{e:loss}
  \sum_{t} \| \hat \Phi_t - \Phi(I_t) \|^2.
\end{equation}
In order to do so, we modify~\cref{e:nerf} to generate,
in addition to  a color image $\hat I_t$, a feature image $\hat \Phi_t \in \mathbb{R}^{C\times H\times W}$.
This is obtained by modifying the range of the function $c=(c_\text{rgb},c_\Phi)$ to be $\mathbb{R}^{3+C}$, where $C$ is the number of feature channels.
We call the pair $(\sigma,c_\Phi)$ a \emph{neural feature field} to distinguish it from the neural radiance field $(\sigma,c_\text{rgb})$ typically used for view synthesis.

In the context of neural networks, this approach can be understood as a 2D-teacher-3D-student model.
The teacher is the feature network $\Phi$, which is defined in image space, and the student is the network implementing the function $c_\Phi$, defined in 3D world space.
This is illustrated in \cref{fig:method}.
The final training loss for the student network is simply the sum of the image reconstruction loss and the feature reconstruction loss weighed by a factor $\lambda$:
\begin{equation}
\label{eq:final-loss}
\sum_{t} \| \hat I_t - I_t \|^2 + \lambda \| \hat \Phi_t - \Phi(I_t) \|^2.
\end{equation}

The key benefits of this approach are twofold.
First, knowledge from the teacher network is \emph{distilled} into the student network in a manner that correctly reflects the 3D geometry of the scene, which has a smoothing effect and helps to regularize feature prediction.
As we show later, this results in higher quality features that are more consistent across viewpoints.
Second, distilling features of general-purpose feature extractors pre-trained on large external datasets\,---\,with or without supervision\,---\,brings open-world knowledge into the 3D representation, which is otherwise scene-specific and lacks semantic understanding.

\subsection{Distillation with advanced NeRF architectures}%
\label{s:advanced}

In \acronym, we are free to implement the neural field $(\sigma, c)$ in any of the many variants that have been proposed in the literature.
In this paper, we showcase the approach on two different scenarios: simple statics scenes, as typically handled by NeRF (presented before), and egocentric videos that are significantly more complex, for which standard neural rendering models are insufficient.
Again, we stress that many other variants would also apply.

The challenge of egocentric videos is that they contain a mixture of static background objects, foreground objects that are manipulated by the actor, and body parts of the actor themselves (\eg, hands).
We handle them by adopting NeuralDiff~\cite{tschernezki21neuraldiff}, a NeRF-like architecture that automatically decomposes a dynamic scene into these three components, combining a static field representing the background, a dynamic field (with a dependency on time $t$ in addition to space $X$) representing the foreground, and another dynamic field anchored at the camera representing the actor.
We can adapt NeuralDiff to support \acronym simply by considering a feature prediction head in addition to the color and density heads for \emph{each} of the three components (MLPs).

\subsection{Applications of \acronym}%
\label{s:applications}

In addition to employing \acronym on top of different neural rendering models, we demonstrate its versatility by considering various downstream applications: 2D object retrieval, 3D object segmentation, 3D scene editing,
and amodal segmentation.
For all these tasks and for ease of evaluation, we assume that a 2D region is provided as a query for a single given frame $I_t$.
As a particular use case for providing queries, one can think of the user introducing an object of interest which can be then localized, \eg, across a video.
However, we note that providing such annotations is not strictly necessary and one could also consider direct clustering of the distilled features to obtain segmentations of objects without manual input, as shown in~\cite{amir2021deep,melas-kyriazi22deep}.

\paragraph{2D object retrieval.}

Given a collection of images, and given any object from a single reference frame, we would like to find all the occurrences of the same object in the rest of the collection, despite significant viewpoint changes, occlusions, and various dynamic effects.
In particular, given a region $R_t \subset \Omega$ of the image containing a fully or partially visible object at time $t$, or even just a patch, we pool a feature descriptor as the mean of the region's features:
\begin{equation}\label{e:retrieved-region}
  \Phi(I_t)^\text{avg}_{R_t} = \frac{1}{|R_t|} \sum_{u \in R_t} \Phi(I_t)_u.
\end{equation}
To localize the object in another image $I_{t'}, t'\!\not=\!t$, we return as matching region $\tilde R_{t'}$ the set of pixels whose features are sufficiently close to the mean descriptor according to a threshold $\tau$:
$$
\tilde R_{t'}
=
\left \{
  u \in \Omega:
  \| \eta({\Phi(I_q)_u}) - \eta ({\Phi(I_t)^\text{avg}_{R_t}}) \| \leq \tau
\right \},
$$
where $\eta(a) = a/\|a\|$ normalizes the input vector $a$.

Performance on this task directly depends on the quality of the matched features.
Despite the 2D character of this task, the above equations are directly applicable to \acronym, by simply replacing $\Phi(I_t)$ with the distilled feature map $\hat \Phi_t$ obtained \emph{after rendering} the 3D features back to the $t$-th view, as explained above.
Hereby, we denote the \emph{distilled} mean feature vector corresponding to region $R_t$ as $\hat{\Phi}^{\text{avg}}_{tR_t}$

\paragraph{3D object segmentation.}

Since \acronym predicts a 3D field of features, these features can be used directly, \ie \emph{prior} to rendering, to segment a queried object along with its geometry in the 3D space, rather than retrieving it in a series of 2D images.
Formally, given features $\hat{\Phi}^\text{avg}_{tR_t}$ extracted from a single 2D annotation $R_t$ of the object in image $I_t$, we retrieve the 3D region
$
\{X \in \mathbb{R}^3 : \|c_\Phi(X) - \hat{\Phi}^\text{avg}_{tR_t}\| \leq \tau_\Phi \wedge \sigma(X) \geq \tau_\sigma \},
$
where $\tau_\Phi$ and $\tau_\sigma$ denote thresholds for the features of interest and densities respectively.
We note that this application is seamlessly enabled by \acronym, while not possible to address with the 2D teacher network.

\paragraph{Scene editing.}

Instead of extracting a 3D object, we can also suppress it, \ie remove it from the scene.
To achieve this, we can simply set the occupancy $\sigma(X)$ to zero for all 3D points belonging to an object, \ie all points $X$ such that $\|c_\Phi(X) - \hat{\Phi}^{\text{avg}}_{tR_t}\| \leq \tau_\Phi$.
Once again, in our experiments, the object to be removed is identified using a query region in one of the views (object patch or region).

\paragraph{Amodal segmentation.}

We can adjust the querying and retrieval process of our method to handle occlusions in two different ways.
The first corresponds to the 2D object retrieval task; in this case, due to the rendering process, features (just like colors) are ``blocked'' from reaching the camera if they are occluded, for example by the actor in egocentric videos or by other objects.
However, our approach makes it possible to also \emph{see through occluders}, by disabling the occupancies $\sigma$ for regions of the 3D space that contain features dissimilar to the query descriptor (\cref{e:retrieved-region}).
In practice, this amounts to rendering the 3D features after obtaining a segmentation of the object in 3D, as described above.
In this manner, it is possible to obtain a mask of the full extent of the object, as if occluders are removed, which is often referred to as amodal segmentation~\cite{li2016amodal}.
\section{Experiments}%
\label{s:experiments}

\begin{table*}[t]
\centering
\small
\setlength{\tabcolsep}{5pt}
\begin{tabular}{lccccccccccc@{\hskip 1pt}c}
  \toprule
  Method & S01 & S02 & S03 & S04 & S05 & S06 & S07 & S08 & S09 & S10 & Average & ~(abs~gain)   \\
  \midrule
  DINO~\cite{caron21emerging} [ViT-B/8] & 75.75 & 57.25 & 56.46 & 63.11 & 70.56 & 65.81 & 52.28 & 78.28 & 58.19 & 65.79 & 64.35 & \\
  \acronym (DINO) & 83.64 & 67.19 & 69.21 & 80.23 & 78.17 & 77.57 & 64.32 & 83.85 & 76.24 & 82.17 & \textbf{76.26} & \improve{(+11.91)} \\
  \midrule
  DINO~\cite{caron21emerging} [ViT-B/16]  & 77.37 & 53.21 & 48.91 & 57.44 & 68.32 & 60.39 & 40.39 & 74.07 & 53.22 & 62.19 & 60.15 & \\
  \acronym (DINO) & 88.61 & 66.99 & 69.90 & 87.02 & 78.66 & 78.97 & 70.57 & 85.17 & 77.59 & 84.93 & \textbf{78.84} & \improve{(+18.69)} \\
  \midrule
  MoCo-v3~\cite{chen2021empirical} [ViT-B/16] & 70.73 & 54.02 & 48.02 & 52.89 & 67.18 & 57.34 & 43.54 & 73.45 & 47.85 & 60.12 & 57.51 \\
  \acronym (MoCo-v3) & 86.67 & 68.95 & 68.53 & 82.93 & 75.74 & 78.00 & 65.63 & 83.58 & 68.26 & 83.21 & \textbf{76.15} & \improve{(+18.64)}\\
  \midrule
  DeiT~\cite{touvron2020deit} [ViT-B/16] & 55.27 & 40.78 & 38.02 & 42.76 & 54.01 & 51.70 & 37.72 & 61.53 & 40.88 & 52.48 & 47.51 \\
  \acronym (DeiT) & 86.02 & 62.47 & 66.69 & 81.22 & 72.93 & 77.88 & 61.63 & 83.73 & 69.59 & 83.12 & \textbf{74.53} & \improve{(+26.82)}\\
  \bottomrule
\end{tabular}
\caption{\textbf{2D object retrieval.}
We compare the features learned by our approach (NeuralDiff-\acronym) with the 2D teacher features on the task of retrieving 2D objects for 10 scenes of the EPIC-KITCHENS dataset.
We consider features from three self-supervised models, two flavors of DINO~\cite{caron21emerging} and MoCo-v3~\cite{chen2021empirical}, and a supervised one, DeiT~\cite{touvron2020deit}.
We report per scene mean average precision (mAP) results and the overall Average.}%
\label{tab:retrieval}
\end{table*}

\begin{figure*}
\centering
\includegraphics[width=0.95\linewidth]{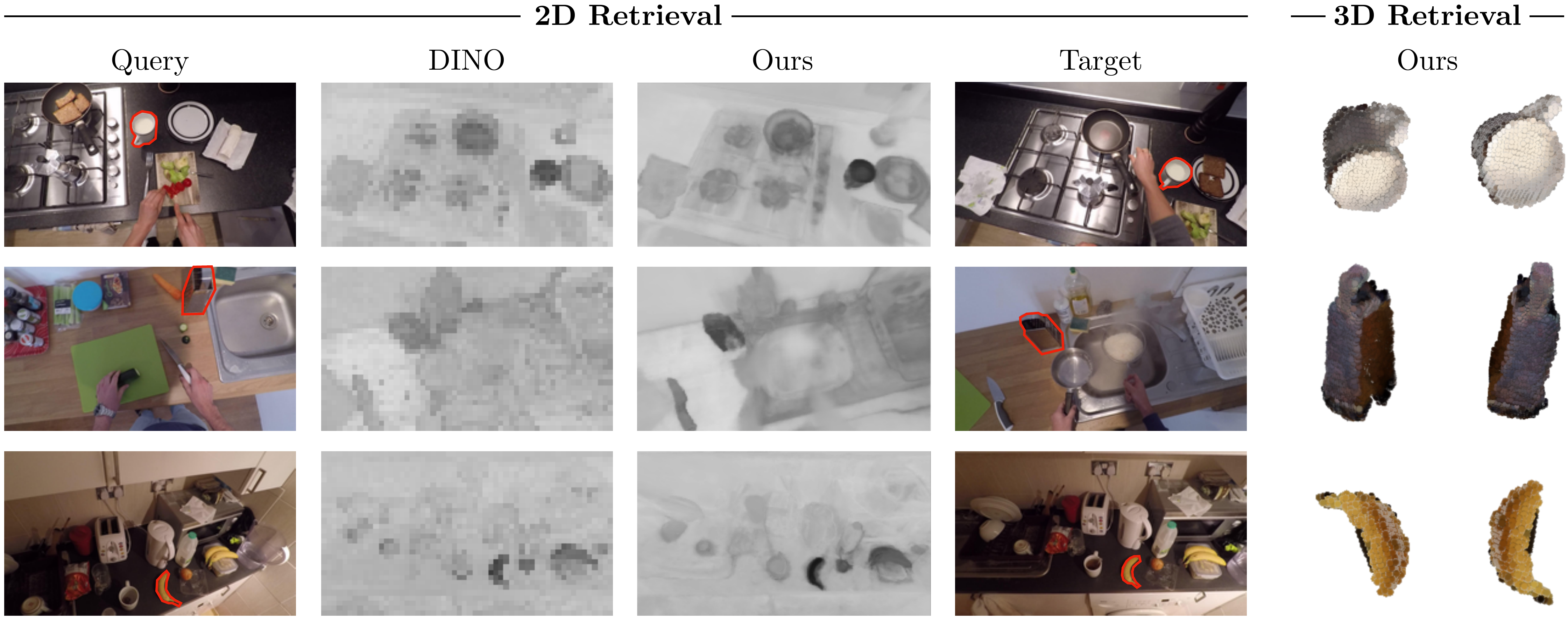}
\caption{\textbf{Retrieving (segmenting) objects in 2D and 3D.}
Given a feature descriptor obtained by pooling features from a given region (Query) in a reference frame, we retrieve similar regions in another frame (Target) of a video sequence.
This can be achieved with either features from a teacher network (DINO) or features learned by our model (NeuralDiff-\acronym). We show that \acronym features are less affected by viewpoint dependent changes such as reflectance, as can be seen for the grater, which has a non-Lambertian surface.
Additionally, our model can compute the densities and colors of 3D features for a given 2D query, which allows us to extract the full 3D extent of objects (seen as point clouds on the right).}%
\label{fig:retrieval-2d-3d}
\end{figure*}
\begin{figure}[t!]
\centering
\footnotesize
\addtolength{\tabcolsep}{-6pt}
\begin{tabular}{cc}
\includegraphics[width=0.99\columnwidth]{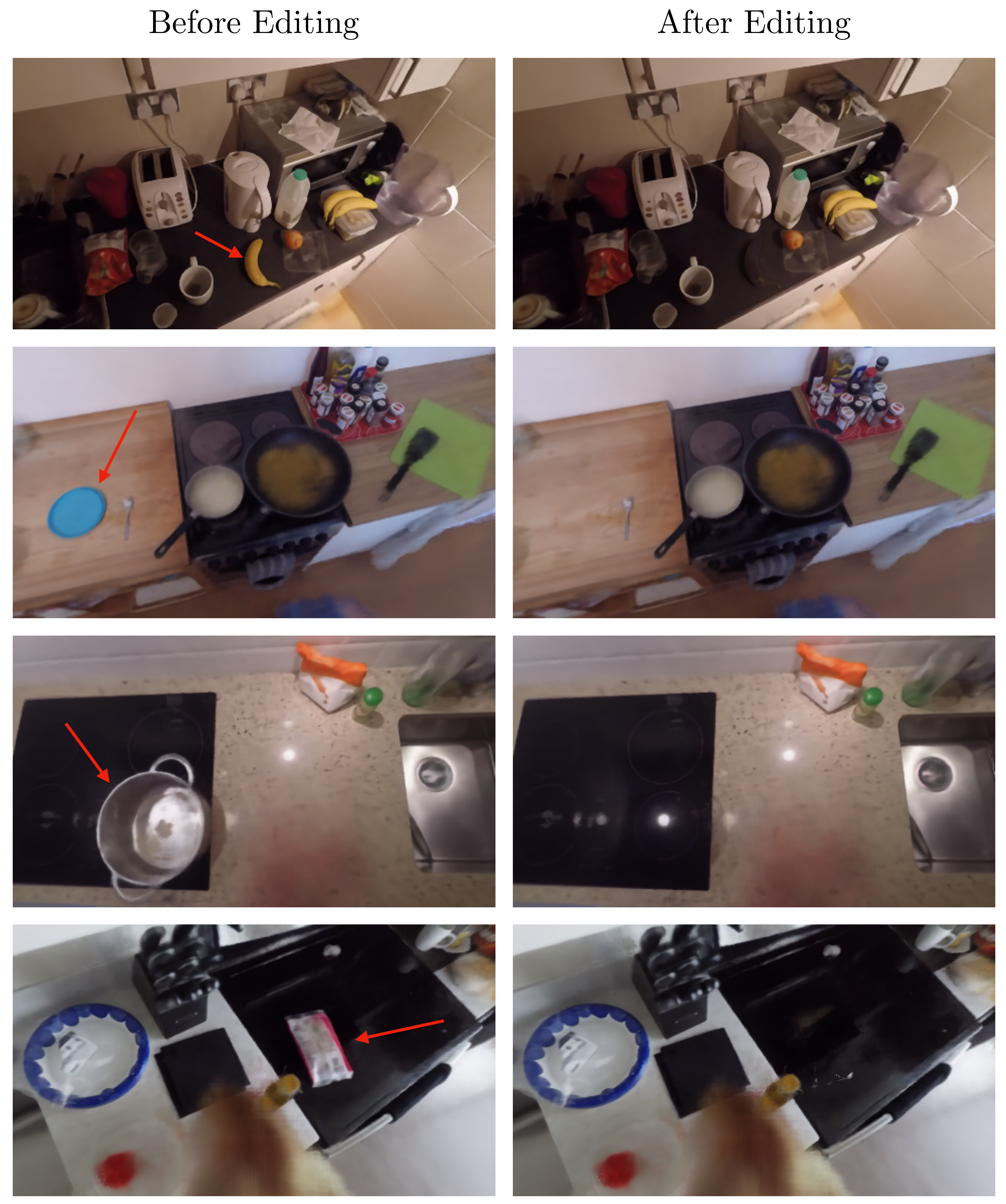}
\end{tabular}
\addtolength{\tabcolsep}{6pt}
\caption{\textbf{Scene editing.}
Our approach allows us to edit a scene in 3D given 2D queries.
Given a 2D segment and its corresponding feature vector, we extract a 3D region of matching features, suppress its occupancies and render the view without the object, \ie removing the banana (first row), lid (second row), pot (third row) and package (fourth row).}%
\label{fig:scene_editing}
\end{figure}

\begin{figure}[t!]
\centering
\includegraphics[width=0.99\linewidth]{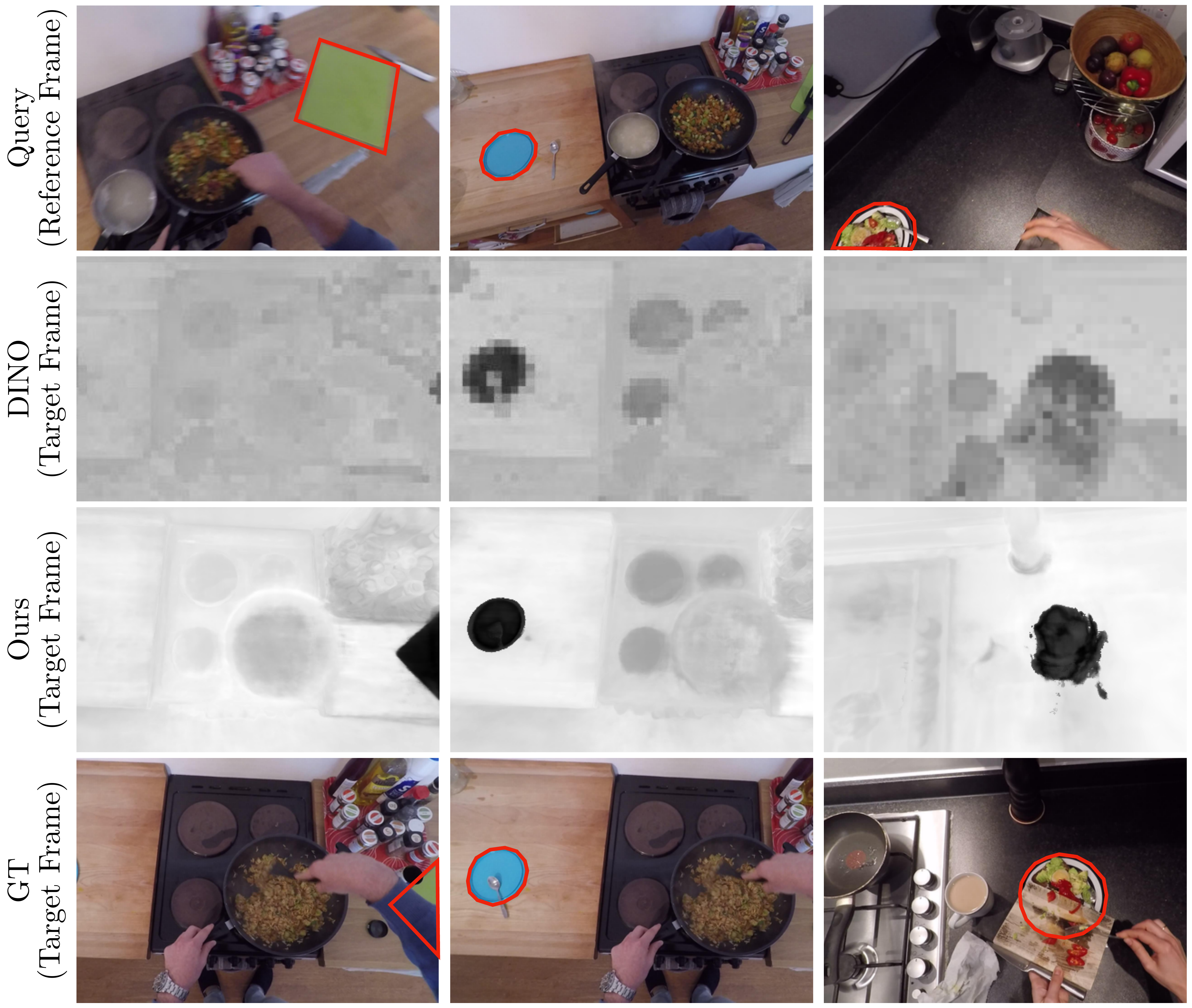}
\caption{
\textbf{Amodal segmentation.}
We compare NeuralDiff-\acronym-distilled features to 2D DINO features for the task of amodal segmentation, \ie segmenting objects through occlusions.
Given a query in a reference frame, \acronym allows us to retrieve the whole object in a target frame despite occluders, by comparing features in 3D and suppressing the occupancies of dissimilar regions prior to rendering.}%
\label{fig:amodal}
\end{figure}

\begin{figure*}[!t]
\centering
\includegraphics[width=0.92\linewidth]{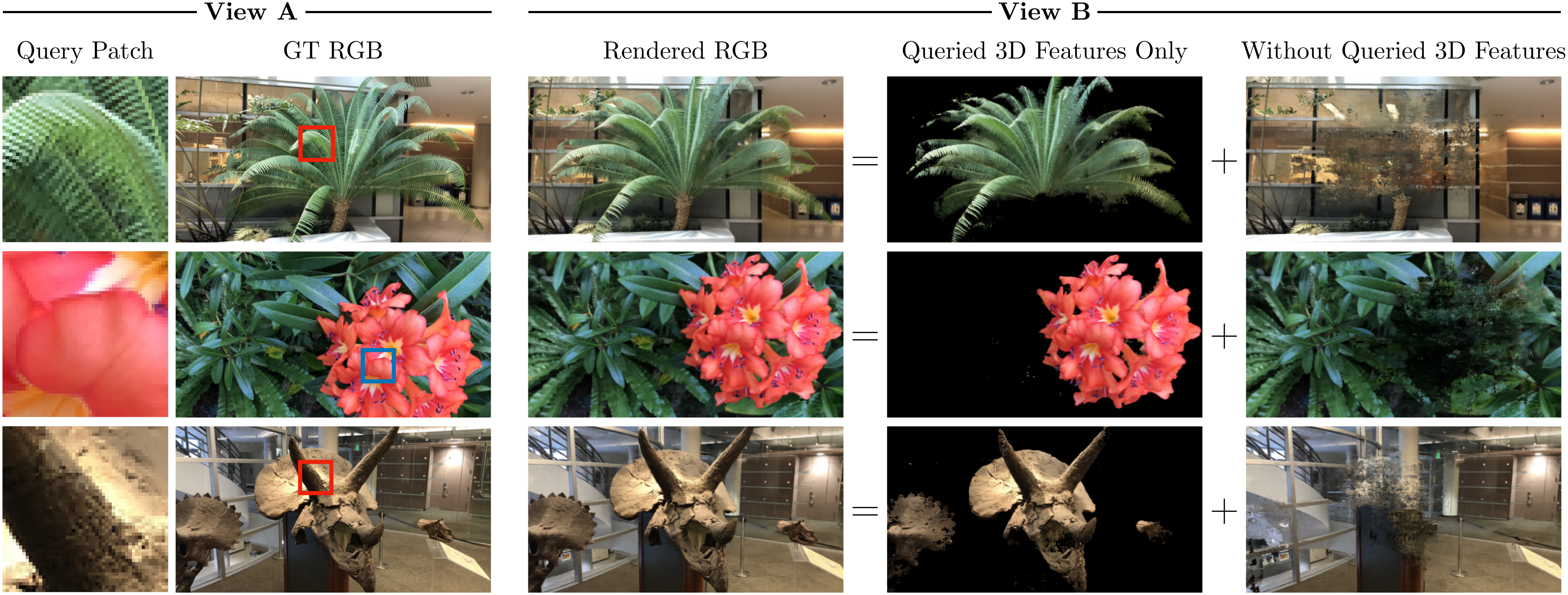}
\vspace{-0.5em}
\caption{\textbf{Scene editing.}
Given a query patch from one (unseen) view, NeRF-\acronym renders an image from another (unseen) view while separating foreground from background by matching  the fused 3D features to the query patch features.
These results also highlight the close relationship to the concurrent work of Kobayashi \etal~\cite{kobayashi2022decomposing}.}%
\label{fig:scene-edit-nerf}
\vspace{-0.5em}
\end{figure*}

\begin{figure}[!t]
\centering
\includegraphics[width=1.0\linewidth]{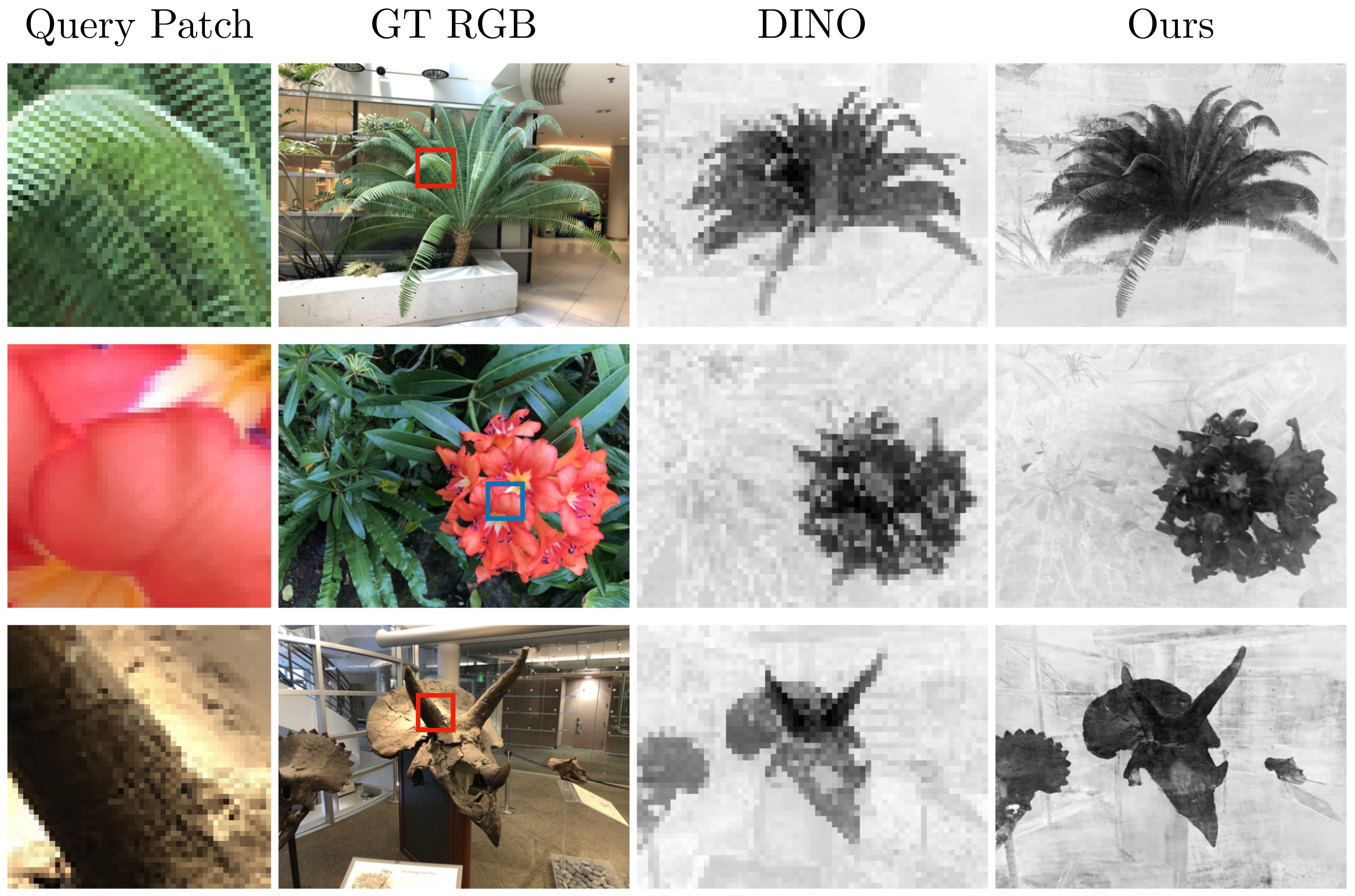}
\caption{\textbf{2D object retrieval.} We calculate feature distance maps with DINO and with our model (NeRF-\acronym) for unseen views from three LLFF scenes.
Our model predicts features for these views through its 3D representation.
}\label{fig:nerf-distill}
\end{figure}

In this section, we evaluate the features produced by \acronym for the tasks introduced in \cref{s:applications} for static and dynamic scenes.
\Cref{sec:setup} gives the experimental details and \cref{sec:results} reports the results for the different tasks.
\Cref{sec:limitations} discusses limitations of our approach.

\subsection{Experimental setup}%
\label{sec:setup}

\paragraph{Datasets.}

We consider scenes from the LLFF dataset~\cite{mildenhall2019llff} and a subset of the EPIC-KITCHENS dataset~\cite{damen18epic}.
The former contains images of static scenes while the latter contains egocentric videos of people cooking in different kitchens, and interacting with a large number of different objects, such as food or kitchen utensils.
For the former, we implement \acronym on top of the vanilla NeRF~\cite{mildenhall20nerf:} architecture, and for the latter on top of the more complex NeuralDiff~\cite{tschernezki21neuraldiff} architecture, as described next.

\paragraph{NeRF.}

For the experiments on the LLFF dataset~\cite{mildenhall2019llff}, we use the NeRF PyTorch implementation~\cite{lin2020nerfpytorch} with the default hyperparameters.
We adapt the architecture with an additional feature prediction head, consisting of a single linear layer with \texttt{tanh} as activation function.
We use the pre-trained models supplied with the implementation and continue training for 5k iterations, freezing all but the feature prediction head for the first 1k iterations.
The weight for the feature distillation loss $\lambda=0.001$.
The features are rendered similarly to pixel colors as described in \cref{s:neuralfusion}.

\paragraph{NeuralDiff.}

We build on the model proposed in~\cite{tschernezki21neuraldiff} for the experiments on EPIC-KITCHENS\@.
We extend the three-stream architecture with feature prediction heads (followed by \texttt{tanh}), one for each component (static, dynamic, actor).
The model is first trained for RGB reconstruction (10 epochs with 20k iterations each and a batch size of 1024).
Training for a single scene takes approximately 24 hours on an NVIDIA Tesla P40.
We then finetune the model to distill the pre-computed teacher features for 20 epochs, 500 iterations each, with the same batch size (approx.~2 hours) and again freezing all but the feature prediction heads for the first 1k steps
(training from scratch yields similar results, but is slower).
We down-sample images to $480 \times 270$ pixels and upscale the 2D features with nearest neighbour interpolation.
We set $\lambda=1.0$ for the feature distillation loss.
The models are trained using Adam optimization~\cite{KingmaB14}, an initial learning rate of $5 \times 10^{-4}$ and a cosine annealing schedule~\cite{loshchilov2017sgdr}.

\paragraph{2D teacher features.}

We consider four transformer-based feature extractors:
DINO~\cite{caron21emerging} with patch size 8 and 16, MoCo-v3~\cite{chen2021empirical} and DeiT~\cite{touvron2020deit}.
DINO and MoCo-v3 are self-supervised whereas DeiT is trained with supervision (image labels).
Features on all scenes are pre-computed using the publicly available weights (pre-trained on ImageNet~\cite{russakovsky2015imagenet}), which are not further updated during the distillation process.
The features are then $L^2$-normalized and reduced with PCA to 64 dimensions before distillation.

\paragraph{Evaluation metric.}

Each sequence used for quantitative evaluation in the next section has $K$ objects annotated in $N$ different frames, corresponding to ground truth regions
$
\{R_{kn}\}_{
  1\leq k \leq K,
  1\leq n \leq N}.
$
These annotations are used for evaluation only, and never considered during training. 
Details on the annotation process can be found in the supplementary material.
We consider the task of 2D object retrieval for quantitative evaluation and divide the annotated frames into two non-overlapping sets, a query set $Q$ and a gallery set $G$ ($Q \cup G = \{1, \dots, N\}$).
Each region $R_{kq}$ ($q \in Q$) is in turn used as a query, searching for the corresponding object in each annotated frame from the gallery set.
In order to avoid fixing a threshold $\tau$ as in~\cref{e:retrieved-region}, for each target frame $I_g$ ($g \in G$), pixels $u$ are sorted by increasing distance to the mean feature ${\Phi(I_q)^\text{avg}_{R_{kq}}}$ ($\hat\Phi^\text{avg}_{R_{kq}}$ for the \acronym-distilled features) and labeled as positive if they belong to the ground truth region $R_{kg}$ and as negative otherwise.
The sorted labels are used to compute the Average Precision (AP) $AP_{kqg}$.
The AP values are then averaged across videos, objects and queries to obtain a mean AP (mAP).

\subsection{Results}%
\label{sec:results}

We present our results for the different tasks mentioned in \cref{s:applications}, namely 2D and 3D object retrieval and segmentation, amodal segmentation, and scene editing.

\paragraph{2D object retrieval.}

In \cref{tab:retrieval} we present quantitative evaluation results for scenes of the EPIC-KITCHENS dataset.
We report the mAP value over different queries for each scene, and the average performance over all scenes.
We compare the distilled features learned by NeuralDiff-\acronym to those of the corresponding 2D teacher networks.
We observe that the 2D features alone perform already well on this task, with self-supervised features (DINO, MoCo-v3) surpassing supervised ones (DeiT).
This is likely due to the fact that models trained with self-supervision have better generalization properties~\cite{sariyildiz2021concept}.
When comparing the 2D features with the distilled features, we observe significant improvements across all feature extractors and all scenes.
The smallest increase occurs when distilling the already strong DINO features, resulting in an absolute difference of 11.9 mAP\@.
The potential for improvement is larger when distilling DeiT features and we observe a larger performance gap, with our model reaching an mAP of 
74.5 vs. 47.5.

We also present qualitative results on both EPIC-KITCHENS (\cref{fig:retrieval-2d-3d}) and LLFF (\cref{fig:nerf-distill}), comparing the features learned by NeuralDiff-\acronym and NeRF-\acronym respectively with those directly obtained from a 2D teacher (DINO).
In \cref{fig:retrieval-2d-3d}, we show objects queried in a given frame by selecting 
an object mask, followed by the resulting distance map in \emph{feature} space for a different frame of the same scene.
Overall, we observe that \acronym increases the clarity and correctness of the maps, resulting in sharper boundaries and higher confidence for the target objects.
For example, in \cref{fig:retrieval-2d-3d} (second row), DINO struggles to recognize the grater in the target frame, possibly due to metallic reflections present in the query and a strong change in appearance.
We observe similar results for NeRF-\acronym in \cref{fig:nerf-distill}, where our approach retrieves the whole object from a small user-provided patch, extracting a more detailed as well as complete segmentation of the objects compared to vanilla DINO\@.
In both scenarios, our approach improves over the 2D teacher by encouraging multi-view-consistency, a property then captured by the distilled 3D features.

\paragraph{3D object segmentation.}

Besides retrieving objects in 2D space, our approach also allows to extract the geometry (\eg, as a point cloud) of a queried region, as detailed in \cref{s:applications}.
We can thus obtain segmentations of various objects in 3D, without requiring any 3D labels to train our models.
This is illustrated in \cref{fig:retrieval-2d-3d} (3D Retrieval).
While details are limited due to the 
precision of the model and partiality of the observations, the recovered shape is broadly correct.
We also note that this task lies outside the capabilities of the original teacher network and is only enabled by the fusion of 2D features into the 3D field.

\paragraph{Scene editing.}

\Cref{fig:scene_editing} shows examples of images rendered with NeuralDiff-\acronym before (left) and after (right) editing.
Given a 2D query region, we find its location in 3D by matching features and suppress its occupancies (setting them to zero), thus removing selected objects.
Note that images are correctly `inpainted' under the object because of the holistic scene knowledge implicitly contained in the radiance field.
This is especially true in the case of the EPIC-KITCHENS data, and dynamic scenes in general, as objects appear at different locations for different time steps.
Thus, removing objects results in valid backgrounds, because the background was observed at some point.
In comparison, scene editing in NeRF-\acronym (\cref{fig:scene-edit-nerf}) results in partially hallucinated background, since part of it is occluded for all viewing directions provided during training.

\paragraph{Amodal segmentation.}

\Cref{fig:amodal} shows qualitative results for the task of  amodal segmentation, \ie~segmenting the full extent of an object, including both visible and occluded parts.
For reference, the figure also shows ``ground truth'' segmentations for these objects, but note that these are manually extrapolated in case of occlusions (since the object is not visible).
Owing to its 3D awareness, our model is able to accurately segment, \eg, the cutting board (first column), even though it is barely visible behind the actor's arm.
In comparison, the teacher network (DINO) cannot segment occluded parts, since it is limited to 2D representations.

\subsection{Limitations and ethical considerations}%
\label{sec:limitations}

\acronym inherits some of the limitations of the source features.
For instance, self-supervised features such as DINO tend to group semantically related objects.
In the EPIC-KITCHENS dataset, we have observed this behavior for objects such as fruits and vegetables or the handles of utensils (pans, kettles, \etc), which are often close in feature space.
This might be undesirable in scenarios where a specific object \emph{instance} should be tracked across a video sequence.

Another limitation is the quality of the 3D reconstruction.
Reconstruction can fail catastrophically in some videos.
In general, details of small or thin objects can be difficult to reconstruct, making it impossible to segment some 3D objects even if they are separated correctly by the 2D features.
An example is the cutting board in~\cref{fig:retrieval-2d-3d} because of its thinness and proximity to the underlying table.

Besides general caveats on the reliability of unsupervised machine learning, there do not appear to be significant ethical concerns specific to this project.
EPIC-KITCHENS contains personal data (hands), but was collected with consent, and it is used in a manner compatible with their terms.
\section{Conclusions}%
\label{s:conclusions}

We have presented \acronym, an approach to boost the 3D consistency of 2D image features within sets of images that can be reconstructed in 3D via neural rendering.
We have shown that \acronym works with various neural rendering models and scenarios, including static objects and harder egocentric videos of dynamic scenes. 
Our experiments illustrate the benefit of our approach for the tasks of object retrieval, segmentation and editing.
Future work includes integrating \acronym in the self-supervised process that learns the 2D features in the first place (\eg, DINO) and fusing multiple videos to establish cross-instance correspondences (\eg, by matching similar utensils in different kitchens).

{\small\paragraph{Acknowledgments.}
We are grateful for support by
NAVER LABS,
ERC 2020-CoG-101001212 UNION,
and EPSRC VisualAI EP/T028572/1.
We thank the anonymous reviewers for their feedback that helped to improve our paper.}

{\small\bibliographystyle{ieee_fullname}\bibliography{biblio}}
\clearpage
\newpage\appendix
\section*{Supplemental Material}

\section{{\method with CNNs}}

While the features of vision transformers stand out in several ways compared to convolutional neural networks (CNNs)~\cite{caron21emerging}, we can also use CNN features with our method.  When evaluating the DeepLab CNN with self-supervised weights from MoCo-v3 on the 10 EPIC-KITCHENS scenes (average over all scenes as reported in \cref{tab:retrieval}), we observe an mAP of 29.7 (teacher features).
Distilling these features results in an improvement of about 20 mAP (NeuralDiff-\acronym reaching 48.9 mAP), which is similar to the improvements observed for the transformer architectures, although the absolute performance remains comparatively low.

\section{Foreground segmentation in videos}

\begin{figure}[h]
\centering
\includegraphics[width=0.99\columnwidth]{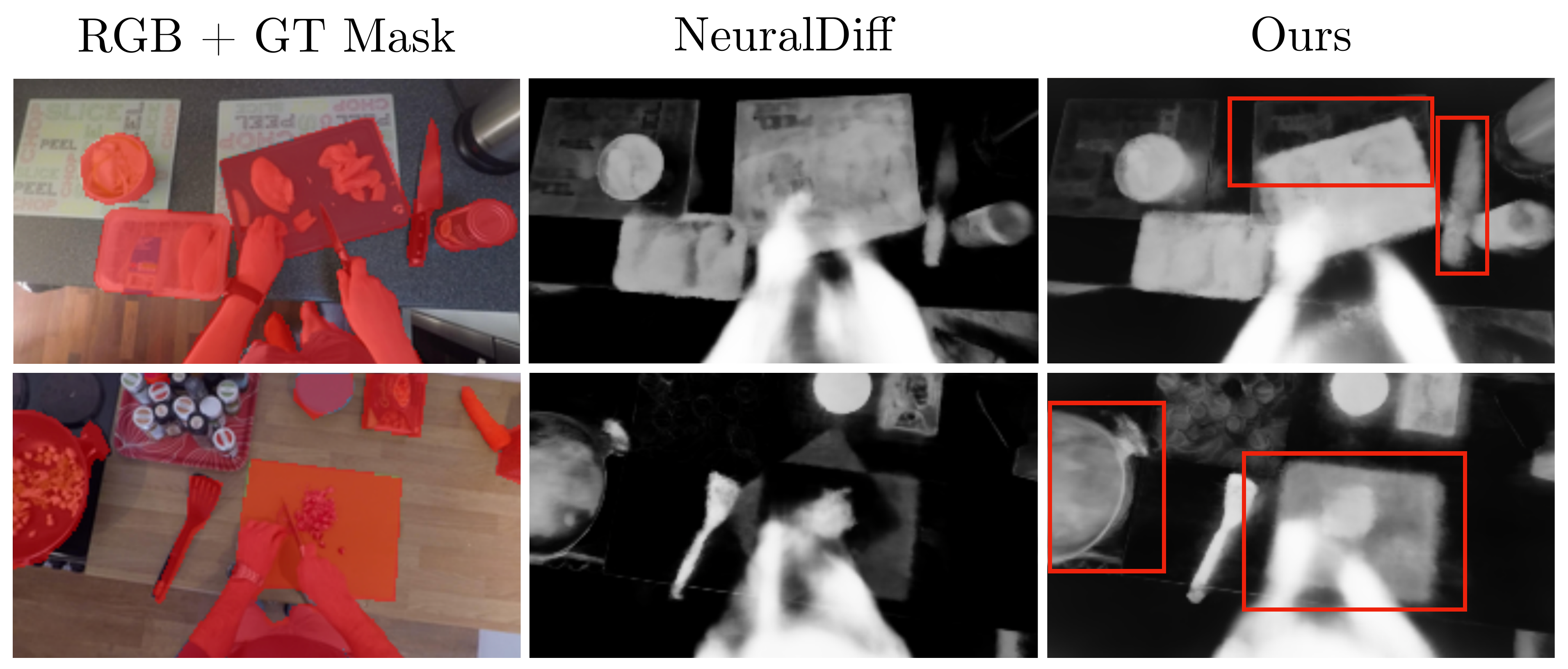}
    \caption{
    \textbf{Foreground segmentation in videos.}
    We show results for the task of foreground object segmentation as described in \cite{tschernezki21neuraldiff} for the original NeuralDiff method~\cite{tschernezki21neuraldiff} and for NeuralDiff-\acronym. We observe that dynamic objects are segmented with higher precision on average when \acronym is combined with NeuralDiff. This is especially true for objects that are moved less frequently during the video (such as the cutting boards in both rows) or objects that are less distinguishable from the background such as the pan in the bottom row (which is black as the hotplate) or the knife in the top row. The red bounding boxes highlight these cases.}
\label{fig:3dv21-segmentation}
\end{figure}

Additionally to the object retrieval results, we also evaluate our model on the segmentation task from \cite{tschernezki21neuraldiff}. In this case, our objective is to segment \textit{all} objects that move at some point during the video sequence and to additionally segment the actor. We use the annotations from \cite{tschernezki21neuraldiff} in our experimental setup, but use all images during training since we are only interested in the performance regarding foreground segmentation and not the photometric reconstruction. We train the standard NeuralDiff model and the same version extended with our method from scratch and evaluate it on the task of foreground object segmentation.
We observe that our model improves the segmentation: NeuralDiff-\acronym results in an mAP of 81.08 while NeuralDiff reaches 76.97 mAP.
We also show qualitative results that highlight the improvements for two scenes in \Cref{fig:3dv21-segmentation}.

\section{Data and protocol for NeuralDiff-\acronym}

\begin{table}[h]
  \centering
  \footnotesize
  \begin{tabular}{ccccccc}
    \toprule
    ID & KID & Query & Val. & Test & $Total$ & Duration \\
    \midrule
    S01 & P01\_01 & 26 & 9 & 13 & 48 & 27 min \\
    S02 & P03\_04 & 27 & 9 & 12 & 48 & 28 min \\
    S03 & P04\_01 & 23 & 8 & 12 & 48 & 19 min \\
    S04 & P05\_01 & 28 & 9 & 14 & 51 & 06 min \\
    S05 & P06\_03 & 31 & 10 & 15 & 56 & 11 min \\
    S06 & P08\_01 & 23 & 8 & 11 & 42 & 10 min \\
    S07 & P09\_02 & 25 & 9 & 12 & 46 & 06 min \\
    S08 & P13\_03 & 25 & 9 & 12 & 46 & 06 min \\
    S09 & P16\_01 & 29 & 9 & 14 & 52 & 20 min \\
    S10 & P21\_01 & 22 & 8 & 11 & 41 & 11 min \\
    \midrule
    - & All & 259 & 88 & 126 & 473 & 144 min \\
    \bottomrule
  \end{tabular}
  \caption{\textbf{Summary of our annotations.}
  Number of annotations per scene for the query and gallery sets as described in \cref{sec:setup}. The gallery is divided into non-overlapping subsets for validation and testing. KID refers to the video ID from the EPIC-KITCHENS dataset. For each scene, we annotate 5 objects.
  }%
  \label{tab:summaryannotations}
\end{table}

We start with the dataset introduced in~\cite{tschernezki21neuraldiff} for all experiments conducted on EPIC-KITCHENS.
As noted in \cref{s:advanced}, their model can handle both static and dynamic objects, separating the background, foreground and actor in three different layers. We extend their dataset by recalculating the COLMAP~\cite{schoenberger16sfm} reconstructions for images using higher resolution. In particular, we downscale the full HD images (from the EPIC-KITCHENS dataset) by a factor of 4 instead of 8 to render images with more details.

For each video sequence, we sample around $|Q| = 26$ frames to build the query set, and $|G| = 21$ frames for the gallery set.
We further divide the gallery into a validation and a test set, of about 9 and 13 frames per scene respectively.
For each sequence, we chose $K=5$ different objects, and annotated all their occurrences in the different frames of the query and gallery sets.
\Cref{tab:summaryannotations} contains more details about the annotations on a per-scene basis.
The objects selected for annotations are arbitrary (\ie they do not belong to any predefined category) and they are  systematically annotated, so that they can be put in correspondence between frames. 
Since the frames used for training \acronym only represent a subset of the entire videos,%
\footnote{The EPIC-Diff dataset~\cite{tschernezki21neuraldiff} only uses about 800 frames per scene, while a scene can contain up to 90000 frames in total.} annotated frames can be separated by hundreds of other frames. The annotations are collected with VIA \cite{dutta2019vgg} in the form of pixel-level masks.

\twocolumn[{%
\renewcommand\twocolumn[1][]{#1}%
\vspace{-1.5em}
\begin{center}
  \centering
  \captionsetup{type=figure}
      \includegraphics[width=0.99\linewidth]{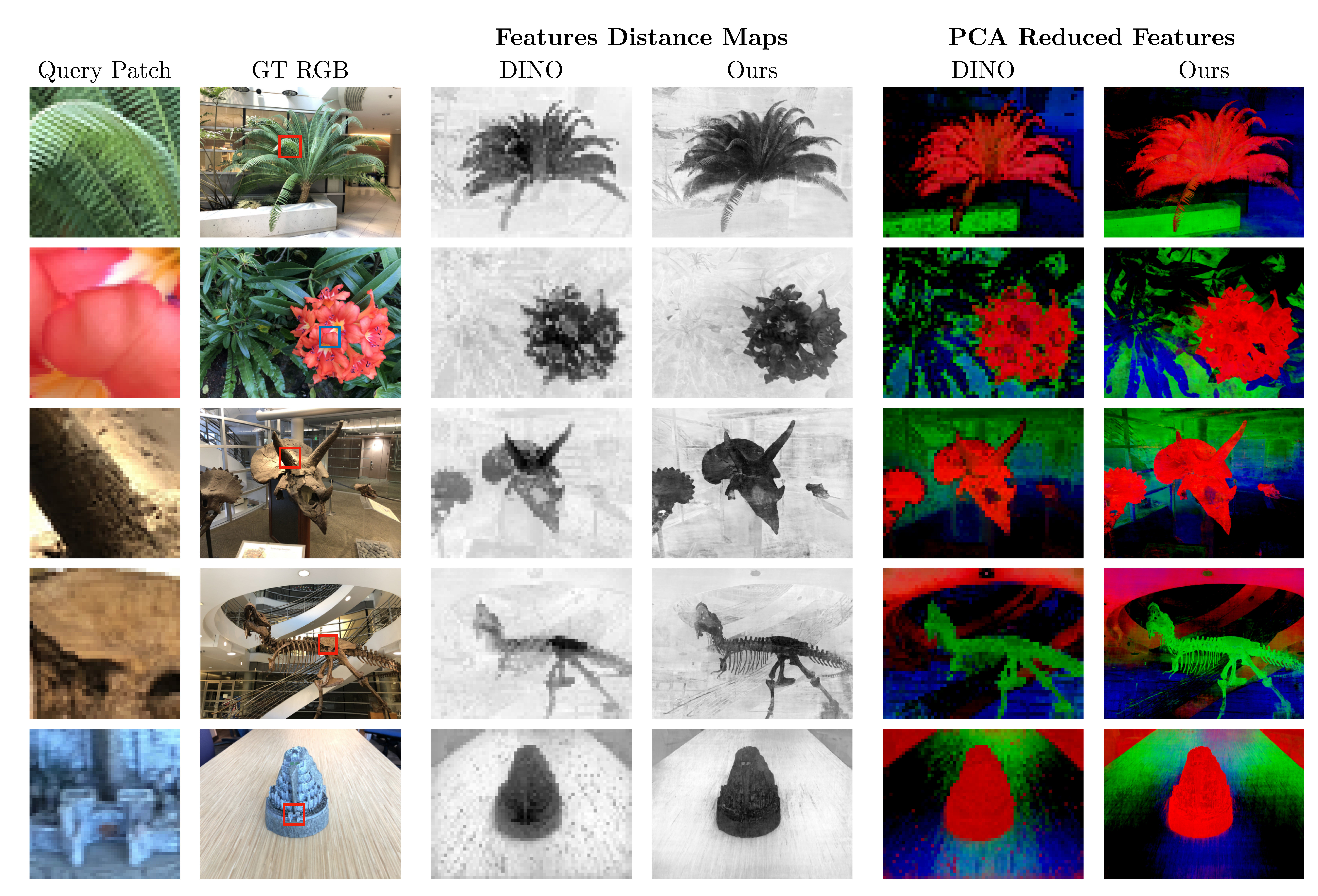}
\captionof{figure}{
\textbf{Retrieval results and PCA reduced features.} Given an unseen view and a query patch from that view, our model renders features from its 3D representation for this patch and retrieves all pixels that correspond to similar features (features distance maps). We also show the rendered features reduced with PCA to three dimensions (RGB channels). While the distance maps for the DINO features focus more on the area around the selected patch, \acronym selects more parts of the entire object more reliably.}
\label{fig:pca_retrieval}
\end{center}%
\vspace{1.5em}
}]

Because objects are manipulated, transformed and partly or entirely occluded in some frames, we need to establish rules on how annotations are carried out.
First, we focus on objects that are actually manipulated during the video, so that matching them cannot be done by mere reconstruction of the static 3D geometry.
Second, only the visible part of each object in each frame is annotated as part of the object mask.
Third, the guideline given to annotators is to mark ``whole objects'', but the latter is sometimes ambiguous and the interpretation is left to the annotator.
For instance, a pan with some food in it can be considered as one, two or more objects.
Usually, it is annotated as a single unit, but we make no effort in providing an exhaustive set of rules to resolve all possible cases.
Note that the annotator's choice is revealed to the retrieval algorithm (\Cref{s:applications}) because the latter is presented with one object annotation each time, that serves as a query for the object retrieval task.
Because of this, the retrieval problem is well-posed.

\section{Additional qualitative results}

In addition to the qualitative results for the three scenes from LLFF visualized in \Cref{fig:nerf-distill}, we show two more scenes and also PCA-reduced features in \cref{fig:pca_retrieval}. We observe that the bias towards the neighborhood around the queried patch is also visible for the t-rex scene (similarly to the dinosaur in row three). Furthermore, the fortress-scene in row five shows less noise regarding the distances of the queried and retrieved features. Finally, the PCA reduced features show higher discriminative power when considering the intensities of the three dimensions (visualized as RGB channels).

\end{document}